\documentclass[10pt,twocolumn,letterpaper]{article}

\usepackage{cvpr}              
\definecolor{cvprblue}{rgb}{0.21,0.49,0.74}
\usepackage[pagebackref,breaklinks,colorlinks,allcolors=cvprblue]{hyperref}

\usepackage{graphicx}
\usepackage{amsmath}
\usepackage{amssymb}
\usepackage{booktabs}
\usepackage{times}
\usepackage{epsfig}
\usepackage{soul}
\usepackage{pifont}
\usepackage{bm}

\newcommand{\model}[1]{\textit{ViFiCon}}
\newcommand{\vifi}[1]{\textit{Vi-Fi}}
\newcommand{\vitagsys}[1]{\textit{ViTag}}
\newcommand{\vitagmodel}[1]{{X-Translator}}

\def\paperID{10} 
\def\confName{CVPR}
\def\confYear{2026}

\title{ViFiCon: Vision and Wireless Association Via \\ Self-Supervised Contrastive Learning}

\author{
Nicholas Meegan$^{*}$, \quad
Hansi Liu$^{*}$, \quad
Bryan Bo Cao$^{\dagger}$, \quad
Abrar Alali$^{\ddagger}$, \\
Kristin Dana$^{*}$, \quad
Marco Gruteser$^{*}$, \quad
Shubham Jain$^{\dagger}$, \quad
Ashwin Ashok$^{\S}$ \\
$^{*}$Rutgers University, \quad
$^{\dagger}$Stony Brook University, \quad
$^{\ddagger}$Saudi Electronic University, \quad
$^{\S}$Georgia State University
}

\begin{document}
\maketitle

\begin{abstract}
   We introduce \model~, a self-supervised contrastive scheme which
   learns a cross-modal association between vision and wireless modalities. Specifically, the system uses 
pedestrian data collected from RGB-D camera footage and WiFi Fine Time Measurements (FTM) from a user's smartphone device. 
Depth data from RGB-D  (vision domain) is inherently linked with an observable pedestrian, but FTM data (wireless domain) is associated only to a smartphone on the network.
We represent temporal sequences from both vision and wireless domains by stacking multi-person depth data sequences within an image representation.  This simplicity allows both scene-wide processing and fewer vision and wireless features, alleviating privacy and energy associated with transmitting IMU data.
To facilitate self-supervised learning, we design a scene-wide synchronization pretext task for our network 
and then employ the learned representation for the downstream multimodal association task.
We show that compared to fully supervised state-of-the-art models, \model~ achieves high performance vision-to-wireless association of \textbf{92.63\%} in 25 frames sliding window fashion (2.5s), 
finding which bounding box corresponds to which smartphone device, 
without hand-labeled association examples for training data. Extensive experimental results demonstrate \model~ applicability in real-world systems when wireless data annotations are scarce.
\end{abstract}

\section{Introduction}

\begin{figure}[t!]
    \begin{center}
    \includegraphics[width=0.98\columnwidth]{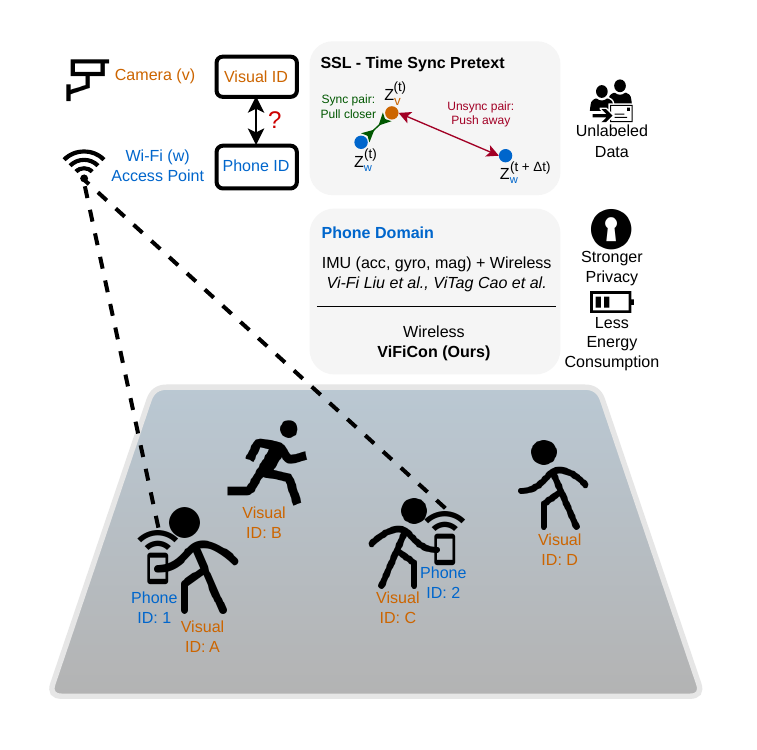}
    \end{center}
    \vspace{-8mm}
    \caption{
     Associating visual observations to WiFi sequences enables novel applications where observed users can be linked to their smart phones. Our approach to learning vision-to-wireless associations does not require annotated ground-truth matches. Instead, a pretext task of  temporal synchronization guides multimodal representation learning for the  downstream task of associating individual bounding boxes to specific smartphones.
    }
    \label{fig:motivation}
\end{figure}

With the proliferation of sensors from multiple modalities,  
cross-domain association for vision and wireless has become a fundamental task in many real-world systems \cite{wang2015visually, cao2018enabling, liuvi, caovitag}. 
The association challenge, i.e.\ correctly matching network addresses with the entities detected in camera images,  enables communication applications such as alerting pedestrians to nearby danger conditions including impassable sidewalks, unusual crowding, high-risk crosswalks, or construction zones.   Additionally, this association supports vision applications in low-light or highly cluttered environments where wireless signals can be used to improve camera-image detection.  

Existing approaches associating information across modalities use visual features with  laser ranging, motion/gait patterns or thermal imagery \cite{spinello2008multimodal, melotti2018multimodal, wang2015visually, lee2022cross}. Among multimodal methods for vision and wireless, RGB-W \cite{alahi2015rgb} fuses camera and received sequence strength (RSS), and \cite{zou2019wifi, fang2020eyefi} fuse camera and WiFi channel state information (CSI). Prior vision and wireless methods are limited by requiring multiple access points (AP). 
Single AP methods for cross-model association of vision and WiFi has emerged in recent state-of-the-art work, including Vi-Fi \cite{liuvi} and ViTag \cite{caovitag}. Our approach \model~ improves efficiency by  
using fewer features overall from both RGB-D and WiFi Fine Time Measurements (FTM), relying only on temporal depth data.
Moreover, unlike Vi-Fi and ViTag, ViFiCon requires no hand-labeled data, improving its practicality for real-world applications.

\noindent \textbf{Challenges.}
To facilitate vision-phone association to a wider range of applications in real world, existing systems encounter several new challenges:  \textbf{\textit{1) Annotations.}} Prior works \cite{liuvi, caovitag}  rely heavily on ground truth labels in order to train their models in a supervised learning manner. However,  time-consuming and labor-expensive annotation hinders the adoption of these systems in novel environments, especially for wireless data which cannot be hand labeled using human visual perception; \textbf{\textit{2) Feature abundance.}} In order to achieve high association accuracy, previous methods consider a wealth of features in phones including accelerations, gyroscope and magnetometer readings as well as wireless data jointly. This assumption can break in a real system since a user may disable  capture and transmission of these data due to privacy \cite{rasnayaka2020your} or energy concerns \cite{kalyankar2022estimation}.

\noindent \textbf{Approach.}
To address these challenges, we  present \model~, a self-supervised contrastive learning model to associate vision and wireless data, circumventing the need of hand-crafted labels. We demonstrate our approach's practical association capability in using only wireless data, alleviating the use of inertial sensors from prior works.

\model~ is among the \textit{first} works to use a shared representation for both vision and wireless modalities  using a Siamese model structure.
%
Multimodal associations  are obtained without hand-labeled data, We construct positive and negative pairings 
across the two modalities through passively collected timestamp information, as inspired by temporal contrastive audio and video synchronization in \cite{korbar2018cooperative}. We create a novel representation of sequences from the two modalities. 
Combining a set of sequences in one shared image learns a global scene-wide synchronization, where each person's FTM and depth in the scene is shown for a window. This guides the downstream one-to-one association task, where we represent each person's depth and FTM data in their own images for sequence matching. 
We train a Siamese convolutional neural network model on the scene-wide synchronization task to embed positive and negative pairs into a joint latent space with a Euclidean distance contrastive loss. 
We can then apply the task downstream without any more training. We show the motivation of \model~ in Figure~\ref{fig:motivation}.

\noindent \textbf{Summary of Contributions}
We summarize our contributions as follows:

\begin{itemize}
    \item We introduce a novel representation of depth and FTM data coined \textit{Vision Depth Sequence (VDS)} and \textit{Wireless Depth Sequence (WDS)} for scene-wide representation without IMU data from phone domain.
    \item We craft a self-supervised learning pretext synchronization task for vision and wireless, using passively collected timestamp information to learn a multimodal joint embedding space without human-labeled data.
    \item \model~ achieves an \textbf{92.63\%} Identity Precision (IDP) in 25-frame temporal window view (2.5s), competitive with fully-supervised approaches while using a simplistic design for a real application.
\end{itemize}
\section{Related Work}

\noindent \textbf{Multimodal Association}
Multimodal association enhances a model's performance by providing additional scene context from shared knowledge across domains. The modalities are fused by mapping data into a shared latent space, where the representations can be compared. Multimodal association has been applied domains such as audio and video \cite{chung2016out, korbar2018cooperative, zheng2023multi, lee2024vmcml}, audio and text \cite{gu2017speech}, and video and wireless \cite{zou2019wifi, masullo2019goes, masullo2020person}. 
We build on prior work that associates vision with wireless sensor data. Wireless data complements visual sensors by overcoming obstructions \cite{adib2013see} and compensating when visual frames are unavailable \cite{zou2019wifi, cao2023vifit, zhang2023layout, zhang2024oostraj, zhang2025out}, hence we leverage WiFi signals with vision to re-identify and tracking in such scenarios. RGB-W \cite{alahi2015rgb} leverages video with RSS from phones to associate boxes with MAC addresses indoors, but requires multiple AP. WiVi and Eye-Fi \cite{zou2019wifi, fang2020eyefi} use CSI along with vision, however, performance degrades under poor lighting and wireless signal interference, and may require multiple AP.
Other techniques use a fusion of FTM, IMU, and vision to perform association and tracking between domains while using one AP and an affinity matrix \cite{liuvi, liu2021lost}, the \vitagmodel~ network \cite{caovitag, cao2022tagging} or a GAN-based approach \cite{liu2023vifi}. We simplify the multimodal association by using vision and wireless sequences mapped into a common space.

\noindent \textbf{Self-Supervised Learning}
Learning without ground truth labels via Self-Supervised Learning (SSL) has been studied in many domains. The core intuition is to learn meaningful features from unlabeled data, usually by a set of pretext tasks. Examples include reconstructing by masking tasks to learn the context, or a similarity latent space by contrastive learning. These tasks have been applied successfully in a wide range of applications in different realms, including learning unimodality in vision \cite{chen2020simple, jing2021self, he2022masked, wei2022masked, Petrovai_2023_WACV}, NLP \cite{devlin2018bert}, audio \cite{manocha2021cdpam}, inertial sensors \cite{xu2021limu} and wireless \cite{liu2021contrastive, yang2022autofi, li2022unsupervised, alloulah2022self, li2022unsupervised, xu2022dual} domains. Though joint domains have been studied, they are mainly limited in vision and language \cite{dai2017contrastive, sarto2023positive, Li_2023_WACV}. In comparison, our work explores the unique research question for real-world multimodal association systems by leveraging contrastive learning to bridge \textbf{vision} and \textbf{wireless} features jointly.
 



\noindent \textbf{WiFi Fine Time Measurements (FTM)}
The FTM protocol introduced in the IEEE 02.11-2016 Standard \cite{80211standard, location2017brings} allows for distance with a margin of error to be collected indoor and outdoor through wireless ranging. Round trip time is calculated between the AP and the connected device, with the range then estimated using propagation speed to bring meter-level precision \cite{ibrahim2018verification}. 


\section{System Overview}
\subsection{Dataset}
Several existing datasets pose limitations to the task of vision-phone association, due to their uni-modality limit such as ImageNet \cite{ImageNet2015}, COCO \cite{lin2014microsoft}, MOT \cite{milan2016mot16} for visual tasks in the vision domain, or IMU data (e.g. HHAR \cite{stisen2015smart}, UCI \cite{reyes2016transition}, Motion Sensen \cite{malekzadeh2019mobile}, Shoaib \cite{shoaib2014fusion}) for the human activity recognition (HAR) task ~\cite{stisen2015smart, malekzadeh2019mobile} in the phone domain. In comparison, \vifi~ \cite{datasetwebsite} provides a \textbf{unique joint} dataset that combines various modalities including vision and phone with IMU and WiFi FTM readings. This comprehensive multimodality offers us a suitable dataset and therefore is chosen for our experiments. We present the necessity of dataset information in this paper and refer readers to the details of data collection in \vifi~ \cite{liuvi}.

From the \vifi~ dataset, sequences in outdoor scenarios are used, including 3 participants and a number of passers-by in the background with at most 11 persons in one frame.

\subsection{Data Pre-Processing}
We formally present the details of data pre-processing in our approach.

\noindent \textbf{Camera Data.}
We follow the main procedure described in \vitagsys~ \cite{caovitag} to tracklets, which are a sequence of one modality's data. We use a time series sequence of bounding boxes (\textit{BBX} to represent a Tracklet from camera data ($T_c$)). A bounding box is defined as: $BBX = [x, y, d, w, h], \quad T_{c} \in \mathbb{R}^{\bm{k} \times \bm{5}}$,
\noindent where the centroid of the bounding box is representedy by $x$ and $y$ while $d$ depth measurement. Bounding box sizes are represented by the width $w$ and height $h$, respectively. We use $k$ to indicate the window's length -- the number of datapoints in the temporal sequence in a window. Unlike prior works \cite{liuvi, caovitag} using all five dimensional camera data, we represent a Tracklet $T_{v} \in \mathbb{R}^{\bm{k} \times \bm{1}}$ in a much simpler form that utilizes only depth information:
A subject's depth in one frame is summarized as a scalar by averaging all the depth values of the region within its bounding box.

\noindent \textbf{Wireless Data.} We use the range of FTM measurements to form the sequence of wireless data, denoted as a Tracklet $T_{w} \in \mathbb{R}^{\bm{k} \times \bm{1}}$.


\noindent \textbf{Multimodal Synchronization.}
Due to various sampling rate in the raw data, we perform synchronization of all the modalities before feeding to the model. Network Time Protocol (NTP) on the devices is used to synchronize the camera and phone data. The camera frame original sampling rate is 30 FPS, while FTM is sampled by a much lower rate at 3-5 Hz. We therefore sub-sample the depth data with 10 FPS to better capture the movement of pedestrians and participants in the scene. To upsample FTM, different from \vitagsys~ \cite{caovitag} where a wireless datapoint finds a camera frame with the closest timestamp, we extract the $k$ nearest larger and smaller timestamps from the FTM data and perform a linear interpolation to align the sequences. This allows better data for the missing FTM datapoints.

\subsection{Vision \& Wireless Depth Sequence Representation}

The depth from the RGB-D camera and FTM information for an individual is a 1-dimensional sequence of values varying over each timestamp. Prior work either deals with a fixed number of pedestrians in a scene a \cite{liuvi}, or processes each tracklet individually \cite{caovitag}, leaving the burden of associating all samples in a scene to an association module. These representation designs pose some limitations that cannot handle association when the number of correspondences $N_{c}$ exceeds the predifined parameter, or the complexity scales linearly by $N_{c}$. To push these limits, we here introduce novel representations coined \textit{\textbf{Vision Depth Sequence (VDS)}} and \textit{\textbf{Wireless Depth Sequence (WDS)}}, formally defined as $D_{v} \in \mathbb{R}^{\bm{H} \times \bm{W}}$ and $D_{w} \in \mathbb{R}^{\bm{H} \times \bm{W}}$
where $H$ and $W$ is the resolution of the image. Specifically, a Tracklet (e.g. $T_{v}$) is mapped to a unique set of gray-scale values in one dimension for each frame that forms a row. We expand the row with some height values represented in one row in a $D_{v}$ image. When multiple samples are present in scene, we stack all their corresponding representations vertically separated by gray padding in rows whose color is determined by the average of the two to clearly define the sequences.

This simple depth image representation enjoys two advantages: First, it allows multiple sequences of walking data to be processed in the same frame, resulting in a global activity view for the scene. Second, finding the number of correspondences $N_{c}$ between the depth and FTM data is trivial in $D$. We summarize the characteristics in comparison to prior works in Table \ref{tab:b}.

\begin{table}[h]
\begin{center}
  \begin{tabular}{cccc}
    \toprule
    Method & No Fixed $N_{c}$ & Granularity & Complexity \\
    \midrule
    \vifi & & $\times$ & \textbf{Scene} & $\textbf{O(1)}$ \\
    \vitagsys & & $\checkmark$ & Subject & $O(N_{c})$ \\
    \textbf{\textit{\model}} & \textbf{$\checkmark$} & \textbf{Scene} & $\textbf{O(1)}$ \\
    \bottomrule
  \end{tabular}
 \caption{Summary of the characteristics of representations. Compared to prior works, Vision Depth Sequence (VDS) and Wireless Depth Sequence (WDS) enjoy the features of both no fixed number of correspondences $N_{c}$ and scene-wise computation, yielding time complexity of processing one scene in $O(1)$.}
 \label{tab:b}
 \end{center}
\end{table}

\begin{figure*}[t]
    \begin{center}
    \includegraphics[width=0.9\textwidth]{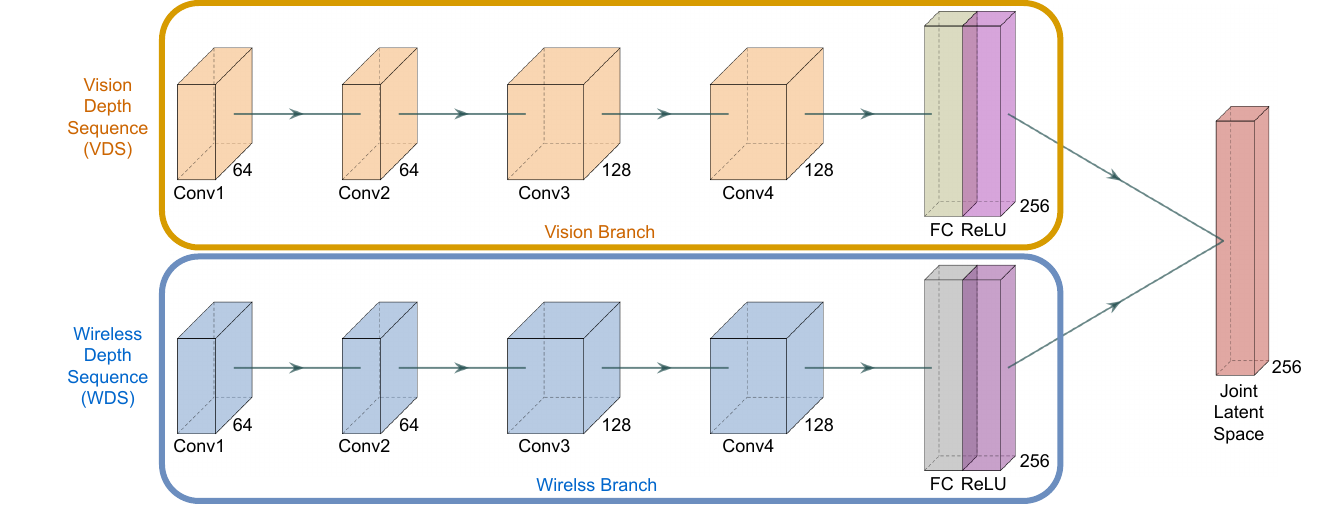}
    \end{center}
    \caption{Network Architecture. Inspired by SimCLR \cite{chen2020simple}, we employ a dual-stream convolutional neural network structure to process the representations of the vision and wireless modalities. Each branch learns independent domain features followed by a shared latent space. Network updates are performed through the contrastive loss. We show that this simplicity is sufficient and beneficial in real world scenarios in terms of efficiency.}
    \label{fig:network_architecture}
\end{figure*}

This representation combines the benefit from both the existing work of \vifi~ \cite{liuvi} and the model \vitagmodel~ in \vitagsys~ \cite{caovitag}. To be specific, \vifi~ handles a fixed number of pedestrians for association ($N_{c}$), while \model~ and \vitagmodel~ do not pose limits on $N_{c}$. However, \vitagmodel~ processes data for each subject in sequence while \model~ and \vifi~ deal with all subjects in one scene simultaneously, resulting in the time complexity of $O(N_{c})$ for the former and $O(1)$ for the latter two, respectively.

\noindent \textbf{VDS $\&$ WDS Preparation} We first split the multimodal data from each sequence into two separate processing steps for depth and FTM ranging data. Using an off-the-shelf ZED detection model, we extract boxes with ID values corresponding to individuals in the scene. We extract depth sequences by taking the average depth across the bounding box for each subject in one frame. The number of returned sequences varies based on the number of individuals in the frames of a window. For wireless, we extract the fixed number of ranging data for each phone. Each sequence is mapped into the Wireless Depth Sequence (WDS) for the pretext task of scene synchronization which will be described in the following sections. In the downstream association task, the sequences are split individually into their own depth sequences.

We note that the order of the sequences does not matter, as the model will pick up on spatial cues to perform matching. In the depth sequence image representation, a variable number of depth sequences may exist in the $k$ frame temporal window. The wireless modality gives an indication at how many users in the scene hold phones, and we use this information for vision to create combinatorics of depth data. One challenge that arises is the fact that some combinatorics may not strictly match the data collected from the wireless domain, including when only participant walking data is selected. 
We threshold for depth values that are not constant for a long period in the window, signifying that an individual has yet to entered or exited the scene.

\section{Methodology}

\subsection{Pretext Task}

\begin{figure*}[h]
    \begin{center}
    \includegraphics[width=0.715\textwidth]{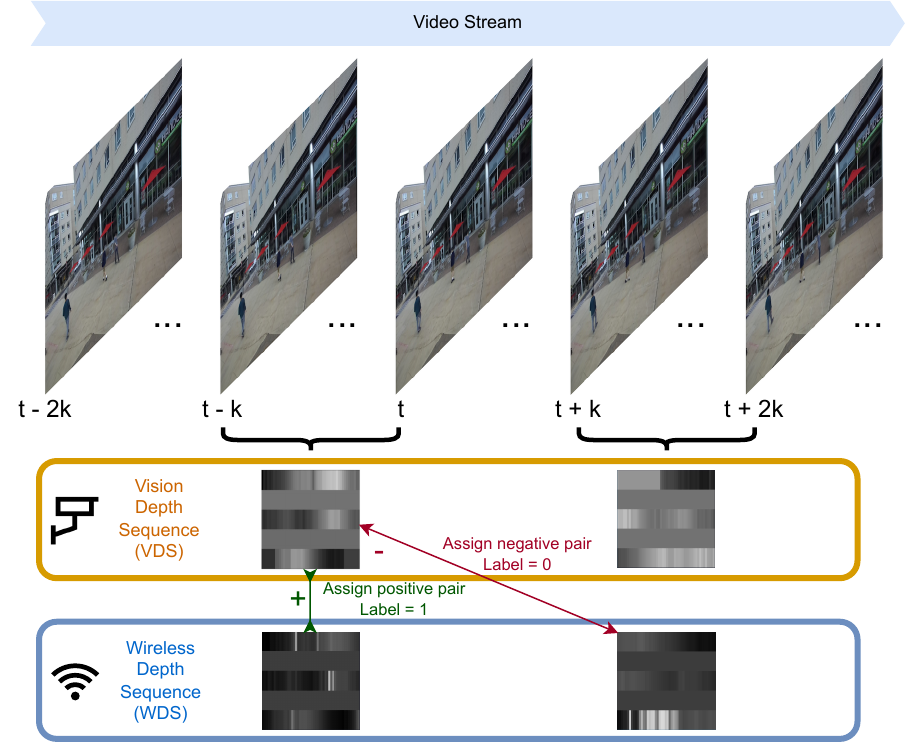}
    \end{center}
    \caption{Self-Supervised Temporal Alignment Pretext Task. For each sequence, we use an off-the-shelf object detector to obtain bounding box information for pedestrians in the scene. We then take the average depth of the bounding box to extract depth sequences for each pedestrian, and create combinations of depth sequences with respect to the fixed number of FTM sequences in the scene if possible. Each image has depth sequences for multiple people in the scene. Temporal alignment or synchronization of these images (pre-text task) can be self-supervised and the resulting representation that can then be used for the association task (downstream task). (Best viewed when zoomed).}
    \label{fig:generation_pairs}
\end{figure*}

Since each datapoint is associated with a timestamp, we consider positive pairs as temporally synchronized Vision and Wireless Depth Sequences , while negative pairs are temporally unsynchronized. Note the label in the association task is the correspondence of a pair of samples between vision and wireless domains (e.g. indicating whether they share the same ID). The main intuition of the pretext task is to leverage timestamps in the dataset, which actually act as \textbf{free} correspondence labels. This way, we can exploit data without human annotations for training. 
The workflow to create positive and negative pairs is outlined in Figure~\ref{fig:generation_pairs}.
Specifically, for a window with size $k$, we generate both $VDS$ and $WDS$ from vision and wireless data, respectively. We take two windows starting at $t-k$ and $t+k$ for example, from each we have $VDS^{(t-k)}$, $WDS^{(t-k)}$, $VDS^{(t+k)}$ and $WDS^{(t+k)}$ shown in Figure~\ref{fig:generation_pairs}. Since $VDS^{(t-k)}$, $WDS^{(t-k)}$ are in the same window (sync), the contrastive loss (which will be described in later sections) will pull their latent vectors to be closer. In contrast, $VDS^{(t-k)}$ and $WDS^{(t+k)}$ are in different windows (unsync), therefore our model will push them far apart in the joint latent space. After training by the time synchronized pretext task, our model will learn a good cross-modal similarity space for the downstream association task.


\subsection{Model Architecture}

Real-world application deployment poses challenges in acquiring computation resources often limited by end users' budget. Albeit the success of Vision Transformer \cite{dosovitskiy2020image, liu2021swin, Mo_2023_WACV} in learning discriminative representations in recent years, this often comes at the cost of high-end computation power and large datasets \cite{pmlr-v139-touvron21a, park2022vision}. In addition, exploring various architectures (such as CNN vs. RNN vs. transformer) can be a promising future work. However, we focus on solving real-world problems towards applications by contrastive learning, and we show that a CNN-backbone is sufficient and efficient to achieve a high association performance using a consumer computer shown in Section \ref{sec:exp}. A summary of the differences is shown in Table \ref{tab:modelcomp}. Our network design is inspired by SimCLR  \cite{chen2020simple}.

\begin{table}[h]
\begin{center}
  \begin{tabular}{cccc}
    \toprule
    Model & FLOPs (G) $\downarrow$ & \#Params (M) $\downarrow$ \\
    \midrule
    ViT-B & 33.03 & 86.86 \\
    ViT-L & 116.68 & 304.72 \\
    \model~ & \textbf{0.75} & \textbf{25.70} \\
    \hline
  \end{tabular}
 \caption{Comparison between Transformer (ViT) and CNN (\model~) as backbones. \model~ enjoys the efficiency advantage for a real application.}
 \label{tab:modelcomp}
 \end{center}
\end{table}




\subsection{Loss Function}
We create our self-supervised contrastive learning model as a set of two identical subnetworks to process the Vision and Wireless Depth Sequence representations. The result is a Dual-stream CNN responsible for fusing the vision and wireless data together in a joint latent space, shown in Figure ~\ref{fig:network_architecture}. 
Determining the embedding distance of each of the vectors in the latent space is performed using the Euclidean distance (ED), given by equation \ref{eqn:ED}:


 \begin{equation}
     ED(Z_{v}, Z_{w}) =
     ||Z_{v}-Z_{w}||_{2}
     \label{eqn:ED}
\end{equation}

\noindent where $Z_{v}$ represents the 256-dimensional feature vector embedding $Z_{v} \in \mathbb{R}^{\bm{256}}$ for the vision depth data, and $Z_{w}$ represents the 256-dimensional feature vector embedding $Z_{w} \in \mathbb{R}^{\bm{256}}$ for the wireless FTM data. The loss back-propagated through the model is a standard contrastive loss function using the Euclidean distances of the feature vector embeddings. 
The contrastive loss (CL) is defined by equation \ref{eqn:CL}:

 \begin{equation}
    \footnotesize
     CL = 
     {1\over B} \times (Y\times||Z_{v}-Z_{w}||_{2}^2 + (1-Y)\times \max(M - ||Z_{v}-Z_{w}||_{2}^2, 0))
     \label{eqn:CL}
\end{equation}

\noindent where $Y \in \{0, 1\}$ is the synchronization label for the given pair, $M$ is a margin hyperparameter, and $N$ is the number of samples. In practice we refer to batch size. For positive pairs, a small Euclidean distance value will yield to a small loss, whereas for negative pairs, so long as the Euclidean distance is larger than the margin hyperparameter $M$, the loss will be minimized. 

We train the \model~ model on a set of 30, 3 minute outdoor sequences from the Vi-Fi using the VDS and WDS for global scene synchronization. Each depth sequence consists of a maximum of 3 sequences on both the vision side and wireless side. On the vision side, training is performed using raw bounding box data extracted from the object detector with no hand tweaking and no ground truth information. We test the model 5 different 3 minute outdoor sequences and compare with state-of-the-art fully supervised models \cite{liuvi, caovitag}. We report results of the pretext synchronization task on the test set and the generalization to the downstream association task. 
\model~ is trained for a total of 400 epochs at a learning rate of 0.003 and a margin hyperparameter of 0.2. 



\section{Experiments and Evaluation Setup}
\label{sec:exp}
\begin{figure*}[h]
    \centering
    \centering
    {\includegraphics[width=0.84\textwidth]{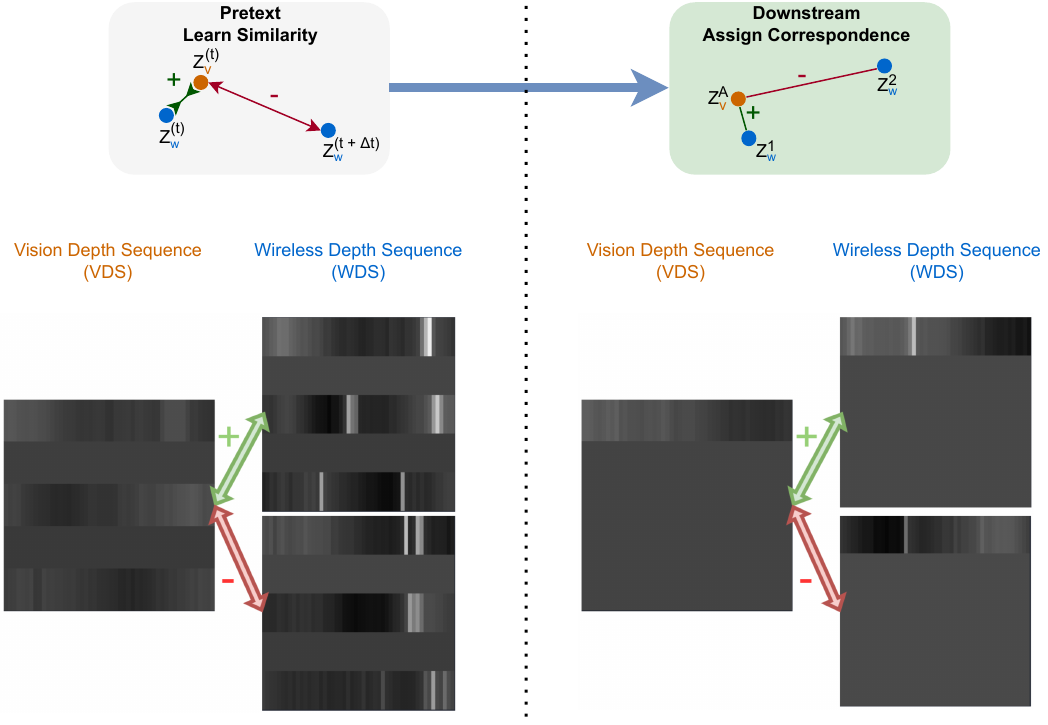}}\label{fig:pretext_downstream_rep}
    \caption{Global scene synchronization (frame-matching) for pretext task and individual association for downstream task. (a) The \textit{Vision Depth Sequence (VDS)} for one frame demonstrates multiple sequences of bounding box depth values (left) and \textit{Wireless Depth Sequence (WDS)} with FTM values (right) for pedestrians in the scene. The pretext global synchronization is a frame-matching problem taking advantage of timestamps without annotations. 
    (b) The downstream association task is an association of a particular \textit{VDS} to a particular FTM \textit{WDS}, using the joint latent space learned from pretext in (a).} 
    
    \label{fig:pretext_downstream}
\end{figure*}

\subsection{Evaluation Metrics}
Following the evaluation protocol in \vifi~ \cite{liuvi} and \vitagsys~ \cite{caovitag}, Our primary evaluation metric is the IDentification Precision \cite{ristani2016performance}, or \textit{IDP}. IDP is defined in equation \ref{eqn:IDP}:

\begin{equation}
    IDP = \frac{IDTP}{IDTP + IDFP}
    \label{eqn:IDP}
\end{equation}

\noindent where IDTP is the identified true positive pairs, and IDFP is the false positive pairs in the latent space. We also provide supplementary metrics of the accuracy and F1 score for evaluating the \model~ model on both the pretext synchronization task and downstream association task

\begin{figure}[h]
    \begin{center}
    \includegraphics[width=0.86\columnwidth]{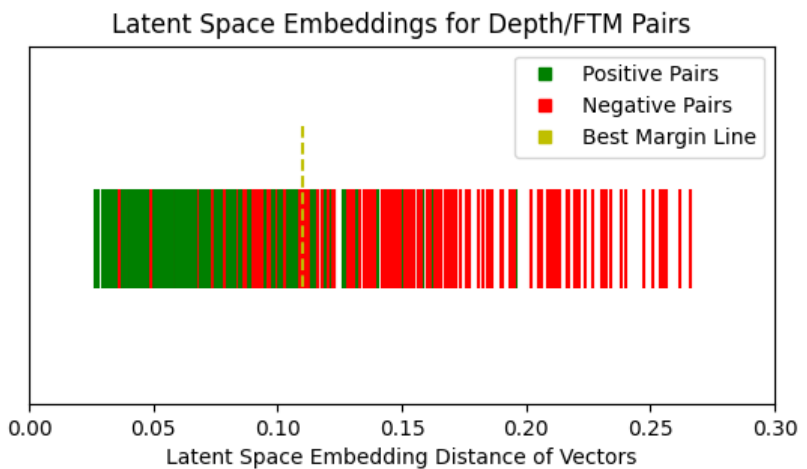}
    \end{center}
    \caption{Sample Latent Space Embedding on downstream association task on test set. We use the true and false positive and negatives to evaluate the model performance. Observe that positive pairings of depth and FTM data (synchronized timestamps) are embedded close, whereas negative pairings (unsynchronized timestamps) have a larger distance.}
    \label{fig:sample_latent}
\end{figure}

After the similarity space has been learned, it will be used for the downstream task of association depicted in Figure \ref{fig:pretext_downstream}. Determining which pairs in the latent space correspond to true positives, false positives, true negatives, and false negatives is done using a margin line. We plot each of the Depth-FTM pairing Euclidean Distances in a latent space representation and then determine the best separation of positive and negative pairings. True positives are located to the left of the margin line, and false positives are to the right. 
A sample view of the latent embedding space including margin line separation is shown in Figure~\ref{fig:sample_latent}.

\subsection{Overall Performance}
\model~ achieves an overall association performance with \textit{IDP} \textbf{$92.63\%$} in the 25-frame (10FPS) sliding window online fashion. The high association with a short window duration of $2.5$ seconds demonstrates our system's practical usage in real-world scenarios.



\noindent \textbf{Performance on Downstream Association Task}
We summarize the results of \model~'s downstream association task performance on the test set in Table \ref{tab:downstream}. We see that the model's performance progressively improves in terms of accuracy (\textit{Acc}) and \textit{F1} score as the temporal window size increases, the best \textit{IDP} score is obtained at a temporal window size of 25 frames. 

\begin{table}[h]
\begin{center}
  \begin{tabular}{cccc}
    \toprule
    Metrics & 10F & \textbf{25F} & 50F \\
    \midrule
    \textit{IDP} (\%) & 84.77 & \textbf{92.63} & 85.33 \\
    \textit{Acc} (\%) & 58.33 & 72.65 & \textbf{78.23} \\
    ~\textit{F1}~ (\%) & 67.10 & 77.38 & \textbf{80.00} \\
    \hline
  \end{tabular}
 \caption{Summary of SSL \model~ performance on Downstream Association Task on Test Set.}
 \label{tab:downstream}
 \end{center}
\end{table}

\begin{figure*}[t]
    \centering
     {\includegraphics[width=0.33\textwidth]{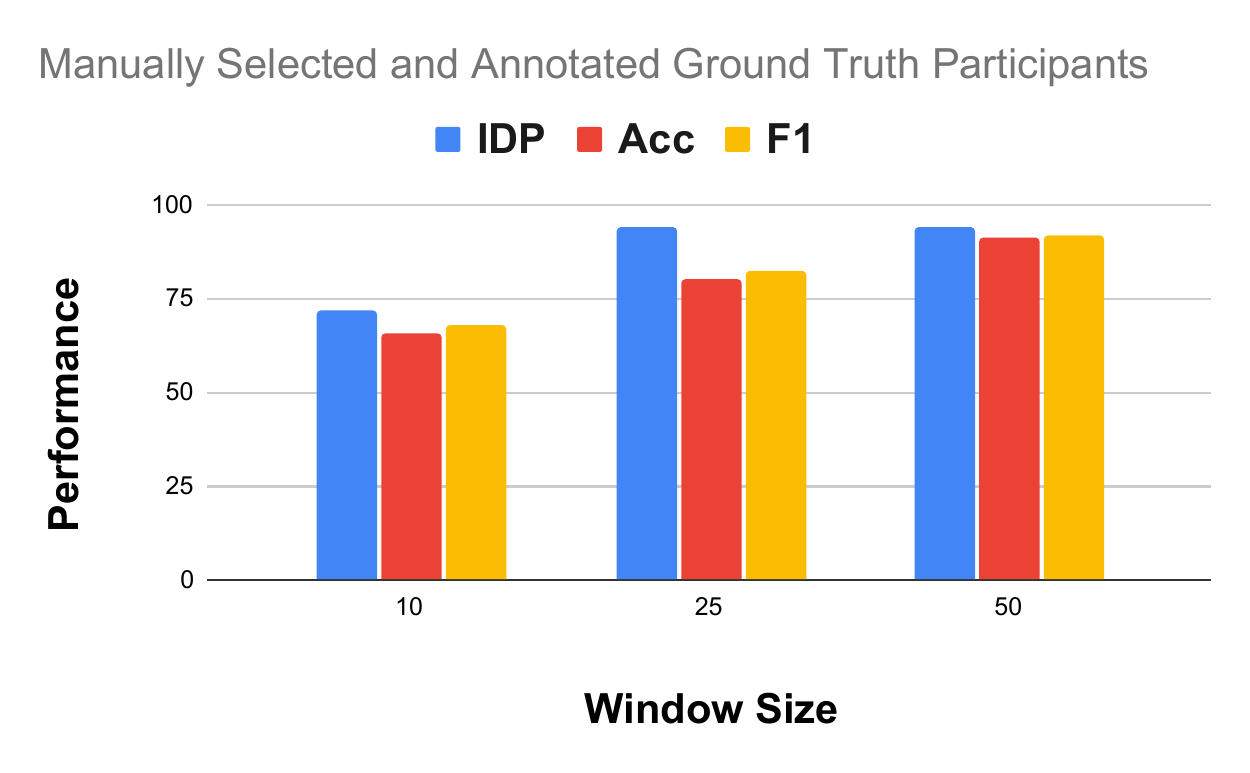}\label{fig:fully_sup}}
    {\includegraphics[width=0.33\textwidth]{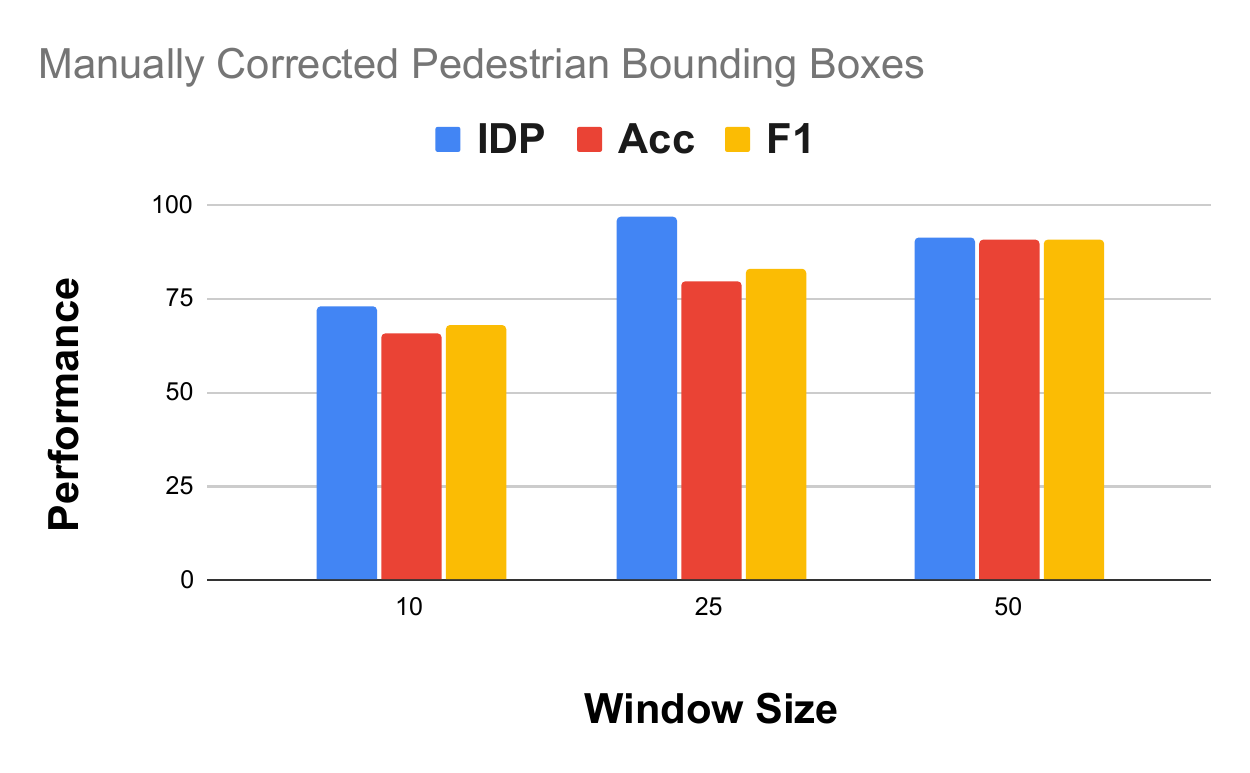}\label{fig:semi_sup}}
    {\includegraphics[width=0.33\textwidth]{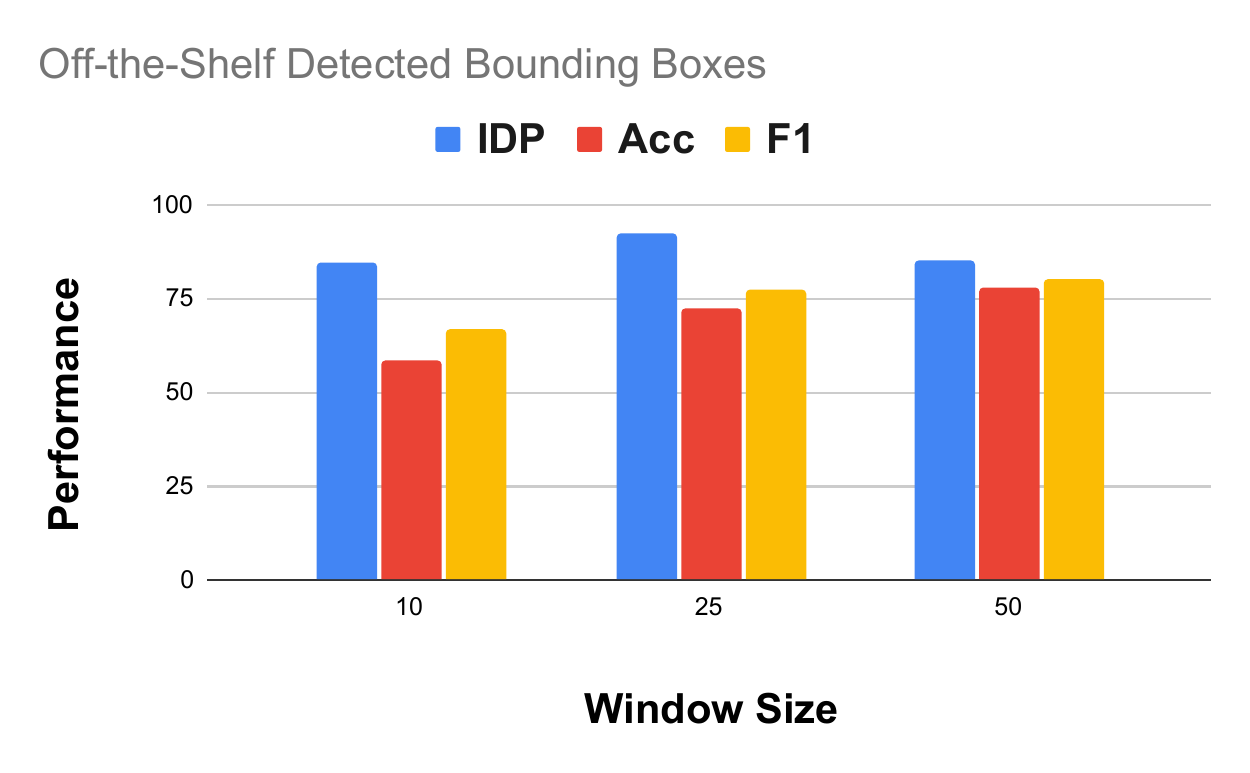}\label{fig:self_sup}}
    
    \caption{ Metrics on manually selected and annotated ground truth (GT), manually corrected pedestrian bounding boxes, and off-the-shelf detector variants for depth data of the \model~ model on the Test Set. 
    In general the manually annotated and selected ground truth has the best performance, though the manually corrected pedestrian bounding boxes and off-the-shelf detector perform comparably due to our thresholding technique.} 
    \label{fig:model_types}
\end{figure*}

\noindent \textbf{Comparison to SOTA models.} We compare our \textit{IDP} score with the \vifi~ ~\cite{liuvi} and \vitagsys~ ~\cite{caovitag} fully supervised deep learning models on the association task. Since only results on 10-frame windows are reported in \vitagsys~, for a fair comparison we hereby use the same $k=10$ for all methods albeit the highest score of \textit{IDP} $92.63\%$ with $k=25$ for \model.
We also include the hand-crafted method of pedestrian dead reckoning (PDR) and procrustes analysis (PA) ~\cite{wang2018pedestrian, krzanowski2000principles, gower1975generalized}. 
We show that by extracting information from the vision domain to obtain the depth sequences, we are able to obtain comparable results without the need for any hand labeling of the dataset. We compare our results from temporal window size 10 with the fully supervised models on the test dataset in Table \ref{tab:comparison}.

\begin{table}[h]
\footnotesize
\begin{center}
  \begin{tabular}{ccccc}
    \toprule
    Model & \textit{PDR+PA} & \vifi~ & \vitagsys~ & \model~ \\
    \midrule
    {Training Set} & - & Labeled & Labeled & Unlabeled \\
    \midrule
    Features & BBX, & BBX, & BBX, & {Depth,} \\
     & IMU, FTM & IMU, FTM & IMU, FTM & {FTM} \\
    
    \midrule
    {\textit{IDP}} & 41.79~\% & 84.97~\% & {87.85~\%} & 84.77\% \\
    \bottomrule
  \end{tabular}
 \caption{Comparison of SSL \model~ performance with state-of-the-art on Downstream Association Task on Test Set as an end use system. \model~ achieves a comparable performance to SOTA deep learning models, but without ground truth annotations as well as fewer features, only requiring depth and FTM data whereas \vifi~ and \vitagsys~ {use IMU, FTM, and bounding boxes (BBX) from the camera data.}}
 \label{tab:comparison}
 \end{center}
\end{table}


\subsection{System Analysis}

\noindent \textbf{Performance on Pretext Synchronization Task}
We record the results of \model~'s pretext synchronization task performance on the test set in Table \ref{tab:pretext}. We see that in general, \model~'s performance improves for all three metrics as the temporal window size increases, with the best results at a large number of frames as more context is provided.

\begin{table}[h]
\begin{center}
  \begin{tabular}{cccc}
    \toprule
    Metrics & 10F & 25F & 50F \\
    \midrule
    \textit{IDP} (\%) & 88.26 & 82.29 & \textbf{95.83} \\
    \textit{Acc} (\%) & 71.04 & 78.65 & \textbf{85.42} \\
    ~\textit{F1}~ (\%) & 75.53 & 79.40 & \textbf{86.79} \\
    \bottomrule
  \end{tabular}
 \caption{Summary of SSL \model~ performance on Pretext Synchronization Task on Test Set.}
 \label{tab:pretext}
 \end{center}
\end{table}

\noindent \textbf{Case Study: Comparing Off-The-Shelf and Manually Corrected Bounding Boxes} We compare our approach with off-the-shelf detections to two different approaches. In the first, we use the ground truth participant bounding boxes for the depth sequence representation on the depth side. In the second, we manually correct bounding boxes produced by the off-the-shelf model such that the IDs are consistent by hand-tweaking. This approach ensures that a depth sequence representation containing all participants holding phones exists in the set of possible combinatorics, but does not guarantee this image will be selected in the dataset generation and train using this image. We provide tabular results of the study on the downstream task in Table~\ref{tab:wivi_variants} for $k=50$ frames, and provide a chart of the IDP for all scenarios across all temporal window sizes in Figure~\ref{fig:model_types}.

\begin{table}[h]
\footnotesize
\begin{center}
  \begin{tabular}{cccc}
    \toprule
    \textbf{Method} & Manually & Manually  & Off-the-Shelf\\
    \textbf{(50 Frames)} & Selected GT & Corrected   & BBoxes   \\
    & & Pedestrian Bboxes  & \cite{ZedCam}\\
    \midrule
    \textit{IDP} (\%) & \textbf{94.00} & 91.33 & 85.33 \\
    \textit{Acc} (\%) & \textbf{91.50} & 90.48 & 78.23 \\
    ~\textit{F1}~ (\%) & \textbf{91.86} & 90.73 & 80.00 \\
    \bottomrule
  \end{tabular}
 \caption{Summary of \model~ performance on Downstream Association Task on Test Set across manually selected and annotated ground truth (GT) participants, manually corrected pedestrian bounding boxes, and off-the-shelf detector variants for depth data with temporal window size $k=50$.}
 \label{tab:wivi_variants}
 \end{center}
\end{table}


\section{Conclusion}
In this work, we propose \model~, a self-supervised contrastive learning approach for vision–wireless association. \model~ leverages \textbf{unlabeled data} and requires substantially fewer phone-side features than existing SOTA methods, while maintaining competitive performance. The former is achieved by learning a visual and wireless similarity space by a pretext synchronization task in a dual-stream CNN backbone, while the latter is enabled by our novel depth sequence image representation that represents a set of depth and wireless sequences in a shared and comparable space. \model~ achieves an \textit{IDP} of \textbf{92.63\%} using a 25-frame (2.5s) online sliding window with only depth and FTM signals on the real-world Vi-Fi dataset, whereas fully supervised methods require per-frame correspondence labels and additional signals such as bounding boxes and IMU data.


Real-world systems poses challenges of scarce ground truth labels, privacy and energy concerns with continnous IMU sensing. We demonstrate that our self-supervised paradigm, combined with the simple yet expressive depth and wireless representations in \model~, addresses labeling scarcity, privacy risks, and energy constraints in a unified framework. By removing reliance on identity annotations, limiting exposure to sensitive visual information, and avoiding continuous IMU usage, \model~ facilitates practical and scalable multimodal deployment, advancing vision–wireless systems toward real-world applications.

\section{Limitations and Discussions}
While ViFiCon achieves strong performance using only depth and WiFi FTM signals, several limitations remain. Our method relies on accurate pedestrian detection and depth estimation from RGB-D sensors, where detection or depth errors may degrade association quality. In addition, the approach assumes reliable temporal synchronization between camera and wireless devices, and clock drift or missing wireless measurements may affect the synchronization-based pretext task. Future directions also include exploring more expressive multimodal architectures to further improve cross-modal representation learning and association performance.
\section*{Acknowledgment}
 {This research was supported by the National Science Foundation (NSF) NRT NRT-FW-HTF: Socially Cognizant Robotics for a Technology Enhanced Socienty (SOCRATES), No. DGE-2021628 and Grant Nos. CNS-2055520, CNS-1901355, CNS-1901133.}


{
    \small
    \bibliographystyle{ieeenat_fullname}
    \bibliography{bibfile}

@String(IJCV = {Int. J. Comput. Vis.})

@String(ICASSP=	{ICASSP})

@online{datasetwebsite,
    author = "Liu et al.",
    title = "Vi-Fi Multi-modal Dataset",
    url  = "https://sites.google.com/winlab.rutgers.edu/vi-fidataset/home",
    addendum = "(accessed: 05.11.2022)"
}

@online{ZedCam,
    author = "",
    title = "ZED Camera Website (accessed: 04.09.2026)",
    url  = "https://www.stereolabs.com/zed/",
    addendum = "(accessed: 04.09.2026)"
}

@ARTICLE{80211standard, 
author={}, 
journal={{"IEEE Std 802.11-2016 (Revision of IEEE Std 802.11-2012)"}}, 
title={{"IEEE Standard for Information technology--Telecommunications and information exchange between systems Local and metropolitan area networks--Specific requirements - Part 11: Wireless LAN Medium Access Control (MAC) and Physical Layer (PHY) Specifications"}}, 
year={2016}, 
volume={}, 
number={}, 
pages={1-3534}, 
doi={10.1109/IEEESTD.2016.7786995}, 
ISSN={}, 
month={Dec},}

@article{location2017brings,
  title={Brings Wi-Fi Indoor Positioning Capabilities},
  author={Location, Alliance W Wi-Fi CERTIFIED},
  journal={Wi-Fi Alliance.[Online]. Avail-able: https://wi-fi. org/news-events/newsroom/wi-fi-certified-locationbrings-wi-fi-indoor-positioning-capabilities},
  year={2017}
}

@inproceedings{chen2020simple,
  title={A simple framework for contrastive learning of visual representations},
  author={Chen, Ting and Kornblith, Simon and Norouzi, Mohammad and Hinton, Geoffrey},
  booktitle={International conference on machine learning},
  pages={1597--1607},
  year={2020},
  organization={PMLR}
}

@inproceedings{li2022unsupervised,
  title={Unsupervised learning for human sensing using radio signals},
  author={Li, Tianhong and Fan, Lijie and Yuan, Yuan and Katabi, Dina},
  booktitle={Proceedings of the IEEE/CVF Winter Conference on Applications of Computer Vision},
  pages={3288--3297},
  year={2022}
}

@inproceedings{manocha2021cdpam,
  title={CDPAM: Contrastive learning for perceptual audio similarity},
  author={Manocha, Pranay and Jin, Zeyu and Zhang, Richard and Finkelstein, Adam},
  booktitle={ICASSP 2021-2021 IEEE International Conference on Acoustics, Speech and Signal Processing (ICASSP)},
  pages={196--200},
  year={2021},
  organization={IEEE}
}

@article{korbar2018cooperative,
  title={Cooperative learning of audio and video models from self-supervised synchronization},
  author={Korbar, Bruno and Tran, Du and Torresani, Lorenzo},
  journal={Advances in Neural Information Processing Systems},
  volume={31},
  year={2018}
}

@inproceedings{chung2016out,
  title={Out of time: automated lip sync in the wild},
  author={Chung, Joon Son and Zisserman, Andrew},
  booktitle={Asian conference on computer vision},
  pages={251--263},
  year={2016},
  organization={Springer}
}

@inproceedings{gu2017speech,
  title={Speech intention classification with multimodal deep learning},
  author={Gu, Yue and Li, Xinyu and Chen, Shuhong and Zhang, Jianyu and Marsic, Ivan},
  booktitle={Canadian conference on artificial intelligence},
  pages={260--271},
  year={2017},
  organization={Springer}
}

@inproceedings{ristani2016performance,
  title={Performance measures and a data set for multi-target, multi-camera tracking},
  author={Ristani, Ergys and Solera, Francesco and Zou, Roger and Cucchiara, Rita and Tomasi, Carlo},
  booktitle={European conference on computer vision},
  pages={17--35},
  year={2016},
  organization={Springer}
}

@inproceedings{wang2015visually,
  title={Visually fingerprinting humans without face recognition},
  author={Wang, He and Bao, Xuan and Roy Choudhury, Romit and Nelakuditi, Srihari},
  booktitle={Proceedings of the 13th Annual International Conference on Mobile Systems, Applications, and Services},
  pages={345--358},
  year={2015}
}

@inproceedings{spinello2008multimodal,
  title={Multimodal detection and tracking of pedestrians in urban environments with explicit ground plane extraction},
  author={Spinello, Luciano and Triebel, Rudolph and Siegwart, Roland},
  booktitle={2008 IEEE/RSJ International Conference on Intelligent Robots and Systems},
  pages={1823--1829},
  year={2008},
  organization={IEEE}
}

@inproceedings{melotti2018multimodal,
  title={Multimodal CNN pedestrian classification: a study on combining LIDAR and camera data},
  author={Melotti, Gledson and Premebida, Cristiano and Gon{\c{c}}alves, Nuno MM da S and Nunes, Urbano JC and Faria, Diego R},
  booktitle={2018 21st International Conference on Intelligent Transportation Systems (ITSC)},
  pages={3138--3143},
  year={2018},
  organization={IEEE}
}

@inproceedings{alahi2015rgb,
  title={RGB-W: When vision meets wireless},
  author={Alahi, Alexandre and Haque, Albert and Fei-Fei, Li},
  booktitle={Proceedings of the IEEE International Conference on Computer Vision},
  pages={3289--3297},
  year={2015}
}

@inproceedings{zou2019wifi,
  title={WiFi and vision multimodal learning for accurate and robust device-free human activity recognition},
  author={Zou, Han and Yang, Jianfei and Prasanna Das, Hari and Liu, Huihan and Zhou, Yuxun and Spanos, Costas J},
  booktitle={Proceedings of the IEEE/CVF conference on computer vision and pattern recognition workshops},
  pages={0--0},
  year={2019}
}

@article{liuvi,
  title={Vi-Fi: Associating Moving Subjects across Vision and Wireless Sensors},
  author={Liu, Hansi and Alali, Abrar and Ibrahim, Mohamed and Cao, Bryan Bo and Meegan, Nicholas and Li, Hongyu and Gruteser, Marco and Jain, Shubham and Dana, Kristin and Ashok, Ashwin and others}
}

@article{caovitag,
  title={ViTag: Online WiFi Fine Time Measurements Aided Vision-Motion Identity Association in Multi-person Environments},
  author={Cao, Bryan Bo and Alali, Abrar and Liu, Hansi and Meegan, Nicholas and Gruteser, Marco and Dana, Kristin and Ashok, Ashwin and Jain, Shubham}
}

@inproceedings{adib2013see,
  title={See through walls with WiFi!},
  author={Adib, Fadel and Katabi, Dina},
  booktitle={Proceedings of the ACM SIGCOMM 2013 conference on SIGCOMM},
  pages={75--86},
  year={2013}
}

@inproceedings{liu2021lost,
  title={Lost and Found! associating target persons in camera surveillance footage with smartphone identifiers},
  author={Liu, Hansi and Alali, Abrar and Ibrahim, Mohamed and Li, Hongyu and Gruteser, Marco and Jain, Shubham and Dana, Kristin and Ashok, Ashwin and Cheng, Bin and Lu, Hongsheng},
  booktitle={Proceedings of the 19th Annual International Conference on Mobile Systems, Applications, and Services},
  pages={499--500},
  year={2021}
}

@inproceedings{masullo2019goes,
  title={Who goes there? exploiting silhouettes and wearable signals for subject identification in multi-person environments},
  author={Masullo, Alessandro and Burghardt, Tilo and Damen, Dima and Perrett, Toby and Mirmehdi, Majid},
  booktitle={Proceedings of the IEEE/CVF International Conference on Computer Vision Workshops},
  pages={0--0},
  year={2019}
}

@article{masullo2020person,
  title={Person re-id by fusion of video silhouettes and wearable signals for home monitoring applications},
  author={Masullo, Alessandro and Burghardt, Tilo and Damen, Dima and Perrett, Toby and Mirmehdi, Majid},
  journal={Sensors},
  volume={20},
  number={9},
  pages={2576},
  year={2020},
  publisher={Multidisciplinary Digital Publishing Institute}
}

@inproceedings{fang2020eyefi,
  title={Eyefi: Fast human identification through vision and wifi-based trajectory matching},
  author={Fang, Shiwei and Islam, Tamzeed and Munir, Sirajum and Nirjon, Shahriar},
  booktitle={2020 16th International Conference on Distributed Computing in Sensor Systems (DCOSS)},
  pages={59--68},
  year={2020},
  organization={IEEE}
}

@article{gower1975generalized,
  title={Generalized procrustes analysis},
  author={Gower, John C},
  journal={Psychometrika},
  volume={40},
  number={1},
  pages={33--51},
  year={1975},
  publisher={Springer}
}

@book{krzanowski2000principles,
  title={Principles of multivariate analysis},
  author={Krzanowski, Wojtek},
  volume={23},
  year={2000},
  publisher={OUP Oxford}
}

@article{wang2018pedestrian,
  title={Pedestrian dead reckoning based on motion mode recognition using a smartphone},
  author={Wang, Boyuan and Liu, Xuelin and Yu, Baoguo and Jia, Ruicai and Gan, Xingli},
  journal={Sensors},
  volume={18},
  number={6},
  pages={1811},
  year={2018},
  publisher={MDPI}
}

@inproceedings{ibrahim2018verification,
  title={Verification: Accuracy Evaluation of WiFi Fine Time Measurements on an Open Platform},
  author={Ibrahim, Mohamed and Liu, Hansi and Jawahar, Minitha and Nguyen, Viet and Gruteser, Marco and Howard, Richard and Yu, Bo and Bai, Fan},
  booktitle={Proceedings of the 24th Annual International Conference on Mobile Computing and Networking},
  pages={417--427},
  year={2018},
  organization={ACM}
}

@article{cao2018enabling,
  title={Enabling public cameras to talk to the public},
  author={Cao, Siyuan and Wang, He},
  journal={Proceedings of the ACM on Interactive, Mobile, Wearable and Ubiquitous Technologies},
  volume={2},
  number={2},
  pages={1--20},
  year={2018},
  publisher={ACM New York, NY, USA}
}

@inproceedings{stisen2015smart,
  title={Smart devices are different: Assessing and mitigatingmobile sensing heterogeneities for activity recognition},
  author={Stisen, Allan and Blunck, Henrik and Bhattacharya, Sourav and Prentow, Thor Siiger and Kj{\ae}rgaard, Mikkel Baun and Dey, Anind and Sonne, Tobias and Jensen, Mads M{\o}ller},
  booktitle={Proceedings of the 13th ACM conference on embedded networked sensor systems},
  pages={127--140},
  year={2015}
}

@article{reyes2016transition,
  title={Transition-aware human activity recognition using smartphones},
  author={Reyes-Ortiz, Jorge-L and Oneto, Luca and Sam{\`a}, Albert and Parra, Xavier and Anguita, Davide},
  journal={Neurocomputing},
  volume={171},
  pages={754--767},
  year={2016},
  publisher={Elsevier}
}

@inproceedings{malekzadeh2019mobile,
  title={Mobile sensor data anonymization},
  author={Malekzadeh, Mohammad and Clegg, Richard G and Cavallaro, Andrea and Haddadi, Hamed},
  booktitle={Proceedings of the international conference on internet of things design and implementation},
  pages={49--58},
  year={2019}
}

@article{shoaib2014fusion,
  title={Fusion of smartphone motion sensors for physical activity recognition},
  author={Shoaib, Muhammad and Bosch, Stephan and Incel, Ozlem Durmaz and Scholten, Hans and Havinga, Paul JM},
  journal={Sensors},
  volume={14},
  number={6},
  pages={10146--10176},
  year={2014},
  publisher={Multidisciplinary Digital Publishing Institute}
}

@article{ImageNet2015,
Author = {Olga Russakovsky and Jia Deng and Hao Su and Jonathan Krause and Sanjeev Satheesh and Sean Ma and Zhiheng Huang and Andrej Karpathy and Aditya Khosla and Michael Bernstein and Alexander C. Berg and Li Fei-Fei},
Title = {{ImageNet Large Scale Visual Recognition Challenge}},
Year = {2015},
journal   = {International Journal of Computer Vision (IJCV)},
doi = {10.1007/s11263-015-0816-y},
volume={115},
number={3},
pages={211-252}
}

@inproceedings{lin2014microsoft,
  title={Microsoft COCO: Common objects in context},
  author={Lin, Tsung-Yi and Maire, Michael and Belongie, Serge and Hays, James and Perona, Pietro and Ramanan, Deva and Doll{\'a}r, Piotr and Zitnick, C Lawrence},
  booktitle={European conference on computer vision},
  pages={740--755},
  year={2014},
  organization={Springer}
}

@article{milan2016mot16,
  title={MOT16: A benchmark for multi-object tracking},
  author={Milan, Anton and Leal-Taix{\'e}, Laura and Reid, Ian and Roth, Stefan and Schindler, Konrad},
  journal={arXiv preprint arXiv:1603.00831},
  year={2016}
}

@inproceedings{he2022masked,
  title={Masked autoencoders are scalable vision learners},
  author={He, Kaiming and Chen, Xinlei and Xie, Saining and Li, Yanghao and Doll{\'a}r, Piotr and Girshick, Ross},
  booktitle={Proceedings of the IEEE/CVF conference on computer vision and pattern recognition},
  pages={16000--16009},
  year={2022}
}

@inproceedings{wei2022masked,
  title={Masked feature prediction for self-supervised visual pre-training},
  author={Wei, Chen and Fan, Haoqi and Xie, Saining and Wu, Chao-Yuan and Yuille, Alan and Feichtenhofer, Christoph},
  booktitle={Proceedings of the IEEE/CVF Conference on Computer Vision and Pattern Recognition},
  pages={14668--14678},
  year={2022}
}

@article{dai2017contrastive,
  title={Contrastive learning for image captioning},
  author={Dai, Bo and Lin, Dahua},
  journal={Advances in Neural Information Processing Systems},
  volume={30},
  year={2017}
}

@inproceedings{sarto2023positive,
  title={Positive-Augmented Contrastive Learning for Image and Video Captioning Evaluation},
  author={Sarto, Sara and Barraco, Manuele and Cornia, Marcella and Baraldi, Lorenzo and Cucchiara, Rita},
  booktitle={Proceedings of the IEEE/CVF Conference on Computer Vision and Pattern Recognition},
  pages={6914--6924},
  year={2023}
}

@article{devlin2018bert,
  title={Bert: Pre-training of deep bidirectional transformers for language understanding},
  author={Devlin, Jacob and Chang, Ming-Wei and Lee, Kenton and Toutanova, Kristina},
  journal={arXiv preprint arXiv:1810.04805},
  year={2018}
}

@inproceedings{xu2021limu,
  title={Limu-bert: Unleashing the potential of unlabeled data for imu sensing applications},
  author={Xu, Huatao and Zhou, Pengfei and Tan, Rui and Li, Mo and Shen, Guobin},
  booktitle={Proceedings of the 19th ACM Conference on Embedded Networked Sensor Systems},
  pages={220--233},
  year={2021}
}

@inproceedings{liu2021contrastive,
  title={Contrastive self-supervised representation learning for sensing signals from the time-frequency perspective},
  author={Liu, Dongxin and Wang, Tianshi and Liu, Shengzhong and Wang, Ruijie and Yao, Shuochao and Abdelzaher, Tarek},
  booktitle={2021 International Conference on Computer Communications and Networks (ICCCN)},
  pages={1--10},
  year={2021},
  organization={IEEE}
}

@inproceedings{alloulah2022self,
  title={Self-supervised radio-visual representation learning for 6g sensing},
  author={Alloulah, Mohammed and Singh, Akash Deep and Arnold, Maximilian},
  booktitle={ICC 2022-IEEE International Conference on Communications},
  pages={1955--1961},
  year={2022},
  organization={IEEE}
}

@article{xu2022dual,
  title={Dual-Stream Contrastive Learning for Channel State Information Based Human Activity Recognition},
  author={Xu, Ke and Wang, Jiangtao and Zhang, Le and Zhu, Hongyuan and Zheng, Dingchang},
  journal={IEEE Journal of Biomedical and Health Informatics},
  volume={27},
  number={1},
  pages={329--338},
  year={2022},
  publisher={IEEE}
}

@InProceedings{Petrovai_2023_WACV,
    author    = {Petrovai, Andra and Nedevschi, Sergiu},
    title     = {MonoDVPS: A Self-Supervised Monocular Depth Estimation Approach to Depth-Aware Video Panoptic Segmentation},
    booktitle = {Proceedings of the IEEE/CVF Winter Conference on Applications of Computer Vision (WACV)},
    month     = {January},
    year      = {2023},
    pages     = {3077-3086}
}

@InProceedings{Li_2023_WACV,
    author    = {Li, Haopeng and Ke, Qiuhong and Gong, Mingming and Drummond, Tom},
    title     = {Progressive Video Summarization via Multimodal Self-Supervised Learning},
    booktitle = {Proceedings of the IEEE/CVF Winter Conference on Applications of Computer Vision (WACV)},
    month     = {January},
    year      = {2023},
    pages     = {5584-5593}
}

@InProceedings{Mo_2023_WACV,
    author    = {Mo, Shentong and Sun, Zhun and Li, Chao},
    title     = {Multi-Level Contrastive Learning for Self-Supervised Vision Transformers},
    booktitle = {Proceedings of the IEEE/CVF Winter Conference on Applications of Computer Vision (WACV)},
    month     = {January},
    year      = {2023},
    pages     = {2778-2787}
}

@inproceedings{lee2022cross,
  title={Cross-Modality Attention and Multimodal Fusion Transformer for Pedestrian Detection},
  author={Lee, Wei-Yu and Jovanov, Ljubomir and Philips, Wilfried},
  booktitle={European Conference on Computer Vision},
  pages={608--623},
  year={2022},
  organization={Springer}
}

@article{dosovitskiy2020image,
  title={An image is worth 16x16 words: Transformers for image recognition at scale},
  author={Dosovitskiy, Alexey and Beyer, Lucas and Kolesnikov, Alexander and Weissenborn, Dirk and Zhai, Xiaohua and Unterthiner, Thomas and Dehghani, Mostafa and Minderer, Matthias and Heigold, Georg and Gelly, Sylvain and others},
  journal={arXiv preprint arXiv:2010.11929},
  year={2020}
}

@inproceedings{liu2021swin,
  title={Swin transformer: Hierarchical vision transformer using shifted windows},
  author={Liu, Ze and Lin, Yutong and Cao, Yue and Hu, Han and Wei, Yixuan and Zhang, Zheng and Lin, Stephen and Guo, Baining},
  booktitle={Proceedings of the IEEE/CVF international conference on computer vision},
  pages={10012--10022},
  year={2021}
}

@InProceedings{pmlr-v139-touvron21a,
  title =     {Training data-efficient image transformers \&; distillation through attention},
  author =    {Touvron, Hugo and Cord, Matthieu and Douze, Matthijs and Massa, Francisco and Sablayrolles, Alexandre and Jegou, Herve},
  booktitle = {International Conference on Machine Learning},
  pages =     {10347--10357},
  year =      {2021},
  volume =    {139},
  month =     {July}
}

@article{park2022vision,
  title={How do vision transformers work?},
  author={Park, Namuk and Kim, Songkuk},
  journal={arXiv preprint arXiv:2202.06709},
  year={2022}
}

@article{kalyankar2022estimation,
  title={Estimation of Parallel Hybrid Scooter’s Energy Consumption through Real Urban Drive Cycle Using IMU},
  author={Kalyankar-Narwade, Supriya and Chidambaram, Ramesh Kumar and Patil, Sanjay},
  journal={Vehicles},
  volume={4},
  number={1},
  pages={297--313},
  year={2022},
  publisher={MDPI}
}

@inproceedings{rasnayaka2020your,
  title={Your tattletale gait privacy invasiveness of IMU gait data},
  author={Rasnayaka, Sanka and Sim, Terence},
  booktitle={2020 IEEE International Joint Conference on Biometrics (IJCB)},
  pages={1--10},
  year={2020},
  organization={IEEE}
}

@inproceedings{cao2022tagging,
  title={Tagging Vision with Smartphone Identities by Vision2Phone Translation},
  author={Cao, Bryan B and Alali, Abrar and Liu, Hansi and Meegan, Nicholas and Gruteser, Marco and Dana, Kristin and Ashok, Ashwin and Jain, Shubham},
  booktitle={IEEE International Conference on Sensing, Communication, and Networking September 2022},
  year={2022}
}

@inproceedings{cao2023vifit,
  title={Vifit: Reconstructing vision trajectories from imu and wi-fi fine time measurements},
  author={Cao, Bryan Bo and Alali, Abrar and Liu, Hansi and Meegan, Nicholas and Gruteser, Marco and Dana, Kristin and Ashok, Ashwin and Jain, Shubham},
  booktitle={Proceedings of the 3rd ACM MobiCom Workshop on Integrated Sensing and Communications Systems},
  pages={13--18},
  year={2023}
}

@inproceedings{liu2023vifi,
  title={ViFi-Loc: Multi-modal pedestrian localization using GAN with camera-phone correspondences},
  author={Liu, Hansi and Lu, Hongsheng and Data, Kristin and Gruteser, Marco},
  booktitle={Proceedings of the 25th International Conference on Multimodal Interaction},
  pages={661--669},
  year={2023}
}

@inproceedings{zhang2023layout,
  title={Layout sequence prediction from noisy mobile modality},
  author={Zhang, Haichao and Xu, Yi and Lu, Hongsheng and Shimizu, Takayuki and Fu, Yun},
  booktitle={Proceedings of the 31st ACM International Conference on Multimedia},
  pages={3965--3974},
  year={2023}
}

@inproceedings{zhang2024oostraj,
  title={Oostraj: Out-of-sight trajectory prediction with vision-positioning denoising},
  author={Zhang, Haichao and Xu, Yi and Lu, Hongsheng and Shimizu, Takayuki and Fu, Yun},
  booktitle={Proceedings of the IEEE/CVF Conference on Computer Vision and Pattern Recognition},
  pages={14802--14811},
  year={2024}
}

@article{zhang2025out,
  title={Out-of-Sight Trajectories: Tracking, Fusion, and Prediction},
  author={Zhang, Haichao and Xu, Yi and Fu, Yun},
  journal={arXiv preprint arXiv:2509.15219},
  year={2025}
}

@inproceedings{lee2024vmcml,
  title={Vmcml: video and music matching via cross-modality lifting},
  author={Lee, Yi-Shan and Tseng, Wei-Cheng and Wang, Fu-En and Sun, Min},
  booktitle={Proceedings of the IEEE/CVF Conference on Computer Vision and Pattern Recognition},
  pages={2060--2069},
  year={2024}
}

@inproceedings{zheng2023multi,
  title={Multi event localization by audio-visual fusion with omnidirectional camera and microphone array},
  author={Zheng, Wenru and Yoshihashi, Ryota and Kawakami, Rei and Sato, Ikuro and Kanezaki, Asako},
  booktitle={Proceedings of the IEEE/CVF Conference on Computer Vision and Pattern Recognition},
  pages={2566--2574},
  year={2023}
}

@inproceedings{jing2021self,
  title={Self-supervised feature learning by cross-modality and cross-view correspondences},
  author={Jing, Longlong and Zhang, Ling and Tian, Yingli},
  booktitle={Proceedings of the IEEE/CVF conference on computer vision and pattern recognition},
  pages={1581--1591},
  year={2021}
}

@article{yang2022autofi,
  title={AutoFi: Toward automatic Wi-Fi human sensing via geometric self-supervised learning},
  author={Yang, Jianfei and Chen, Xinyan and Zou, Han and Wang, Dazhuo and Xie, Lihua},
  journal={IEEE Internet of Things Journal},
  volume={10},
  number={8},
  pages={7416--7425},
  year={2022},
  publisher={IEEE}
}
}


\end{document}